\documentclass[letterpaper, 10 pt, conference]{inc/ieeeconf}  

\IEEEoverridecommandlockouts                              

\usepackage{graphics} 
\usepackage{graphicx}
\usepackage{amsmath} 
\usepackage{amssymb}  
\usepackage{xcolor}
\usepackage{lipsum}

\usepackage{multirow}
\usepackage{hyperref}
\usepackage{caption}
\usepackage{subcaption}
\usepackage{algpseudocode}
\usepackage{algorithm}
\usepackage{flushend}

\title{\LARGE \bf
Just Round: Quantized Observation Spaces Enable\\Memory Efficient Learning of Dynamic Locomotion
}

\author{Lev Grossman$^{1}$ and Brian Plancher$^{2}$
\thanks{$^{1}$Lev Grossman is with Berkshire Grey, Bedford, MA, USA. {\tt\footnotesize lev.grossman@berkshiregrey.com}}%
\thanks{$^{2}$Brian Plancher is with Barnard College, Columbia University, New York, NY, USA. {\tt\footnotesize bplancher@barnard.edu}}%
}

\begin{document}
\maketitle
\thispagestyle{empty}
\pagestyle{empty}

\begin{abstract}


Deep reinforcement learning (DRL) is one of the most powerful tools for synthesizing complex robotic behaviors. But training DRL models is incredibly compute and memory intensive, requiring large training datasets and replay buffers to achieve performant results. 
This poses a challenge for the next generation of field robots that will need to learn on the edge to adapt to their environment.
In this paper, we begin to address this issue through observation space quantization. 
We evaluate our approach using four simulated robot locomotion tasks and two state-of-the-art DRL algorithms, the on-policy Proximal Policy Optimization (PPO) and off-policy Soft Actor-Critic (SAC) and find that observation space quantization reduces overall memory costs by as much as $4.2\times$ without impacting learning performance.

\end{abstract}

\section{INTRODUCTION}

Deep reinforcement learning (DRL) continues to see increased attention by the robotics community due to its ability to learn complex behaviors in both simulated and real environments. These methods have been successfully applied to a host of robotic tasks including: dexterous manipulation~\cite{Andrychowicz18}, quadrupedal locomotion~\cite{Miki22}, and high-speed drone racing~\cite{Song21}. Despite these successes, DRL remains largely sample inefficient, depending on enormous amounts of training data to learn. As much of this data is kept in replay buffers during training, DRL is extremely memory intensive, limiting the number of computational platforms that can support such operations, and largely confining state-of-the-art model training to the cloud. 
%
For example, OpenAI used 400 compute devices and collected two years worth of experience data per hour in simulation in order to train a physical dexterous robot hand to solve a Rubik's cube~\cite{Akkaya19}. 

At the same time, there has been a recent push to enable learning on physical robot hardware~\cite{Ibarz21}. Using environment experience collected directly on-device and sending this batched data over Ethernet to a workstation, researchers have shown that DRL, and off-policy methods in particular, hold promise in learning robust locomotive policies on physical robot hardware~\cite{Haarnoja18a, Wu22}. 

Unfortunately, for such real-world robots, a virtually unlimited compute and memory budget is unrealistic. Still, many robots, especially those involved in tasks as consequential as search-and-rescue and space exploration~\cite{Gankidi17}, will have to adapt to ever-changing environmental conditions and continue to optimize and update their internal policies over the course of their lifetime~\cite{Thrun95}. As such, to untether these methods from the confines of a lab, data collection and storage must be memory efficient enough to allow for either low-latency networking~\cite{Liu19, Qi21} or for sufficient experience data to be stored on-board edge computing devices~\cite{Fedus20}. As fast and secure updates may not be possible in remote locations or when using bandwidth-constrained or high-latency cloud networks~\cite{Groshev22}, it is imperative to find ways to reduce the overall memory footprint of DRL training.

In this work we begin to address this issue through observation space quantization. We focus on the observation space in particular, as the observations stored in a replay buffer generally consume over 90\% of the total memory footprint of DRL training for state-of-the-art locomotion policies~\cite{Miki22, Haarnoja18}. 
Importantly, unlike simply reducing the number of observations stored in the buffer, which decreases the memory footprint at the cost of reduced learning performance, our quantization scheme is able to reduce memory usage without impacting the training performance. We present experiments across four popular simulated robotic locomotion domains, using two of the most popular DRL algorithms, the on-policy Proximal Policy Optimization (PPO) and off-policy Soft Actor-Critic (SAC), and find that our approach can reduce the memory footprint by as much as 4.2$\times$ without impacting training performance. We open-source our implementation for use by the wider robot learning community.

\section{RELATED WORK}
\textbf{Locomotive Learning and Efficiency:} Deep reinforcement learning (DRL) methods have been successfully applied to a variety of complex simulated~\cite{Lillicrap15, Peng17} and real~\cite{Bogdanovic21} robotic locomotion tasks. Advances in asynchronous algorithms~\cite{Mnih16} and massively parallel simulation~\cite{Rudin21} have enabled faster learning. However, DRL algorithms often remain data intensive and prone to being sample inefficient. This inefficiency can be especially impactful when transferring learned policies from simulation to physical robots~\cite{Tan18} or when learning policies directly on hardware~\cite{Haarnoja18a, Wu22}. To address this, some have found success in learning using substantially reduced buffer sizes and intelligent replay sampling~\cite{Lan22}.
Others have been able to increase the sample efficiency of DRL based locomotion by using more powerful off-policy algorithms~\cite{Haarnoja18}. Still, improving the sample efficiency and thus memory efficiency of DRL remains a major area of interest in the robotic learning community.

\textbf{Compressed and embedded spaces:} Previous work on learning in compressed or embedded spaces has mainly surrounded model-based techniques~\cite{Rybkin21, Fawcett22}, often learning an embedding directly from images~\cite{Watter15, Hafner18}. While model-based learning has been shown to work well in some complex dynamic environments~\cite{Nagabandi18}, model-free methods remain a popular choice in the dynamic locomotion community~\cite{Peng17, Bogdanovic21}. Others have turned to embedded and more descriptive action spaces~\cite{Varin19, Allshire21, Karamcheti21, Wong21} and reduced order models~\cite{Fawcett22} to enable more robust and sample efficient learning. However, these efforts have mainly ignored the impact of observation space compression on model-free learning.

\textbf{Quantization:} Quantization or discretization of deep neural network weights and parameters has seen increased popularity~\cite{Gong14, Hubara16}. These methods are able to achieve competitive performance on classic deep computer vision tasks while reducing the size footprint of the convolutional neural network (CNN) models used. More recently, work has been done to adapt these same techniques to reinforcement learning (RL)~\cite{Krishnan19, Bjork21a}. This work has mainly focused on the quantization of the parameters of the critic or value networks, in order to reduce overall model size and speed up learning. However, little has been done to evaluate the effect of applying quantization to a task's observation space.



\section{LEARNING BACKGROUND} \label{sec:learning_bk}
Reinforcement learning poses problems as Markov decision processes (MDPs), where an MDP is defined by a set of observed and hidden states, $\mathcal{S}$, actions, $\mathcal{A}$, stochastic dynamics, $p(s_{t+1}|s_t, a_t)$, a reward function $r(s,a)$, and a discount factor, $\gamma$. The RL objective is to compute the policy, $\pi^*(s,a)$, that maximizes the expected discounted sum of future rewards, $\mathbb{E}_{s,a}\left(\sum_t \gamma^t r_t \right)$. 
%
We train all learning tasks with two state-of-the-art on- and off-policy reinforcement learning algorithms: Proximal Policy Optimization (PPO) and Soft Actor-Critic (SAC).

\subsection{Proximal Policy Optimization (PPO)}

Proximal Policy Optimization~\cite{Schulman17} is an on-policy, policy gradient~\cite{Sutton00} algorithm that employs an actor-critic framework to learn both the optimal policy, $\pi^*(s,a)$, as well as the optimal value function, $V^*(s)$. Both the policy $\pi_\theta$ and value function $V_\phi$ are parameterized by neural networks with weights $\theta$ and $\phi$ respectively. Similar to Trust Region Policy Optimization (TRPO)~\cite{Schulman15a}, PPO stabilizes policy training by penalizing large policy updates. Additionally, as all updates are computed with samples taken from the current policy, PPO often requires high sample complexity.

During training PPO needs to store the parameters of both its value and policy networks\footnote{Many PPO implementations save space by using one central MLP with two additional single-layer policy and value function model heads. We focus on the standard PPO model approach in this section for clarity.}--usually shallow multilayer perceptrons (MLPs)--as well as its on-policy rollout buffer $\mathcal{D}_k$. $\mathcal{D}_k$ stores $(s, a, r)$ tuples, which are refreshed during each iteration of the algorithm.

By combining both the models and rollout buffer, the general space complexity of PPO can be written as:
\begin{equation}
\texttt{size} = \texttt{sizeof}(\pi_\theta) + \texttt{sizeof}(V_\phi) + \texttt{sizeof}(\mathcal{D}_k).
\end{equation}
Importantly, despite throwing out and re-collecting an entirely new rollout buffer during each iteration, $\mathcal{D}_k$ dominates the overall memory footprint of PPO. 

For example, Miki et al.~\cite{Miki22} uses a student-teacher approach to bridge the sim-to-real gap and enable robust quadrupedal locomotion in rough, wilderness terrain. This process starts by using PPO to train the teacher model--a 3-layer MLP fed by two smaller encoder networks.
Assuming all parameters are stored as 64-bit floats, the three models have a combined size of 
$\sim$1.3MB.
In comparison, the 391-dimensional observation space, 16-dimensional action space, and batch size of 8,300 leads to a rollout buffer of over 27MB of space (assuming 64-bit floats). This means that $\mathcal{D}_k$ accounts for 95\% of the total memory footprint.
%
Furthermore, this size is dominated by the stored observations, which account for 
96\% of the size of $\mathcal{D}_k$ and thus 92\% of the total memory footprint.



\subsection{Soft Actor-Critic (SAC)}

Soft Actor-Critic~\cite{Haarnoja18, Haarnoja18a} is an off-policy RL algorithm that generally extends soft Q-learning (SQL)~\cite{Haarnoja17} and optimizes a ``maximum entropy'' objective, which promotes exploration according to a temperature parameter, $\alpha$:
\begin{equation}
     \mathbb{E}_{(s_t, a_t) \sim \pi}\left[\sum_{t} \gamma^t r(s_t,a_t) + \alpha \mathcal{H}\left(\pi(\cdot | s_t)\right) \right].
\end{equation}

SAC makes a number of improvements on SQL by automatically tuning the temperature parameter $\alpha$ using double Q-learning, similar to the Twin Delayed DDPG (TD3) algorithm~\cite{Fujimoto18}, to correct for overestimation in the Q-function, and learning not only the Q-functions and the policy, but also the value function.

Like PPO, SAC's memory usage comes from its models and off-policy replay buffer. 
Shallow MLPs are once again used to approximate the policy $\pi_\theta$ as well as the two Q-functions $Q_{\phi_1}, Q_{\phi_2}$. Unlike PPO, however, the replay buffer, $\mathcal{D}$, is generally much larger in size as it stores trajectories from every iteration, usually acting as a size limited queue.

By combining the models and replay buffer, the general space complexity of SAC can be written as:
\begin{equation}
\texttt{size} = \texttt{sizeof}(\pi_\theta) + 2*\texttt{sizeof}(Q_\phi) + \texttt{sizeof}(\mathcal{D}).
\end{equation}

As in the case of PPO, $\mathcal{D}$ dominates the memory footprint of SAC. 
For example, Haarnoja et al.~\cite{Haarnoja18} used SAC to train the quadruped Minitaur to walk. The combined number of parameters in their policy and two value networks (ignoring bias terms and again assuming 64-bit floats) was about 2.3MB.
%
With an 112-dimensional observation space, an 8-dimensional action space, and a replay buffer of size 1e6,\footnote{A conservative estimate as they collect 100k-200k samples.} $\mathcal{D}$ consumed about 96.8MB of memory, equating to 97.7\% of the total memory footprint.
%
Again, the size of $\mathcal{D}$ is dominated by the observations, which account for 
92.5\% of its size and 90.4\% of the total memory footprint.



\section{METHOD}

In this work, we focus on one particular type of quantization: numerical rounding. We define the function $\texttt{round}(x, m)$ to be the result of rounding $x \in \mathbb{R}$ to $m \in \mathbb{N}$ decimal places and extend this operation to all real vectors $X \in \mathbb{R}^n$. Thus, rounding an observed state, $\texttt{round}(s, m)$, to $m$ decimal places is the same as $\{\texttt{round}(s_i, m), \; \forall s_i \in s\}$. 

We also make use of a $\texttt{clamp}$ operation with bounds $a, b$:
\begin{equation}
    \texttt{clamp}(s, a, b) = \min (\max (s, a), b),
\end{equation}
This operation is useful for further reducing the number of bits required to express each individual observation $s$ and is a popular tool employed in previous quantization efforts~\cite{Jacob17}.

Combining both operations, we transform an input observation state $s$ into a quantized observation state $s_q$: 
\begin{equation} \label{eq:quant}
    s_q = \texttt{round}(\texttt{clamp}(s, a, b), m).
\end{equation}


In practice, we set $a=-b$ in order to clamp symmetrically around 0. We can thus calculate the number of bits required to encode a quantized output as: 
\begin{equation}
    1 + \lceil \log_2(b) \rceil + \lceil m\log_2(10) \rceil,
\end{equation}
where 1 bit is needed for the sign, $\lceil \log_2(b) \rceil$ bits are needed to encode the integer left of the decimal, and $\lceil m\log_2(10) \rceil$ bits are required to encode $m$ decimal places.

We adapted out quantization scheme to the data from our baseline experiments. In those experiments we found that $\max(|s|) < 128$, which allowed us to set our clamping bounds at $a=-127, b=127$ and use 7 integer bits. Combined with the 1 sign bit and either 4 ($m=1$) or 7 ($m=2$) decimal bits, we were able to quantize the original 64-bit floating-point values down to either 12 or 15 bits, an over $5\times$ and $4\times$ reduction in space respectively.


\section{EXPERIMENTS AND RESULTS}
\begin{figure}[!t]
    \centering
    \includegraphics[width=0.8\columnwidth]{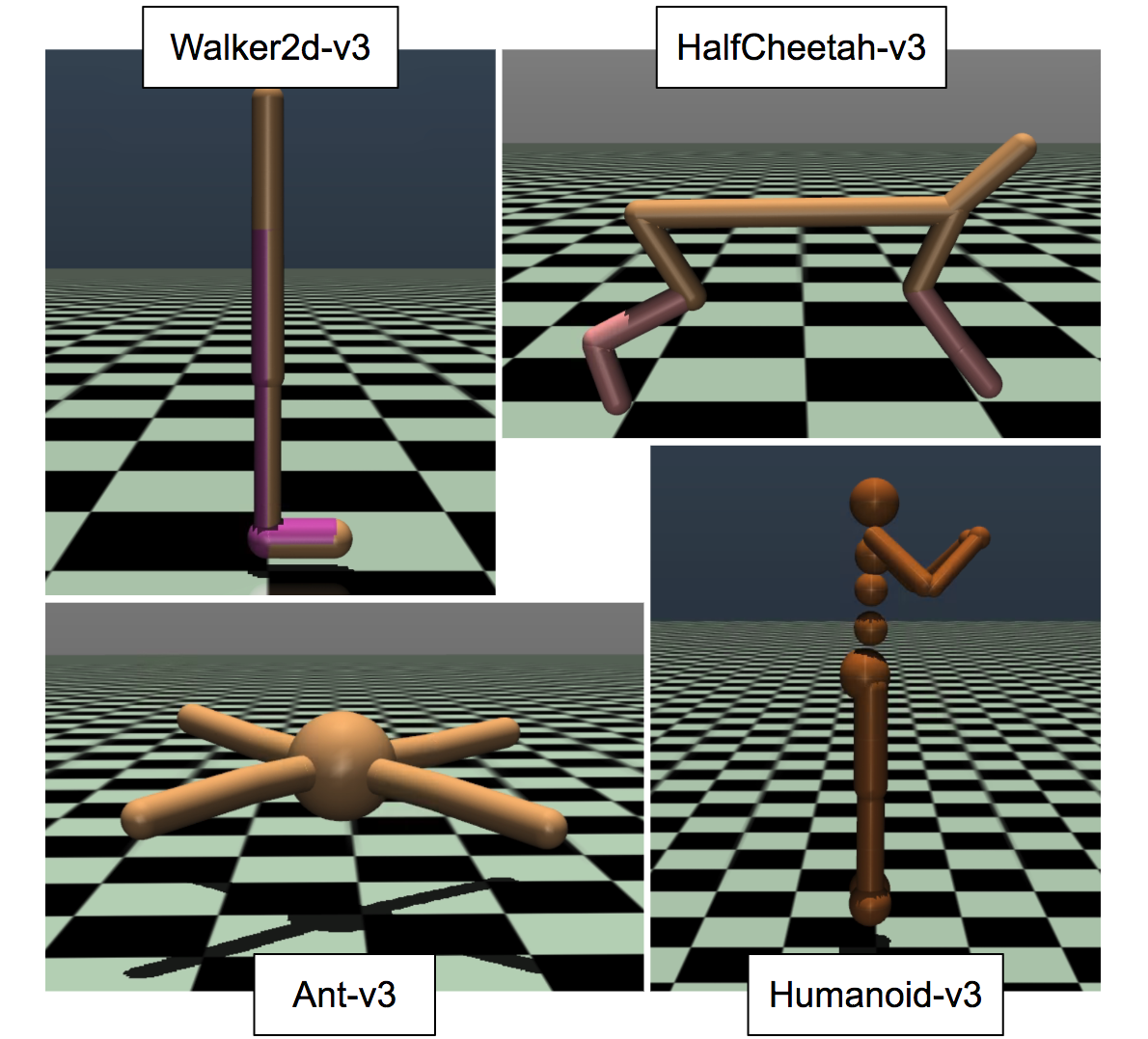}
    \caption{MuJoCo simulated robotic locomotion environments: $
    \texttt{Walker2d-v3}$ (top-left), $
    \texttt{HalfCheetah-v3}$ (top-right), $
    \texttt{Ant-v3}$ (bottom-left), $
    \texttt{Humanoid-v3}$ (bottom-right).}
    \label{fig:environments}
    \vspace{-15pt}
\end{figure}

We evaluate the effectiveness of our quantization framework across four popular, simulated robotic locomotion environments made available by OpenAI Gym~\cite{OpenAIGym}: \texttt{Walker2d-v3}, \texttt{HalfCheetah-v3}, \texttt{Ant-v3}, and \texttt{Humanoid-v3}. Figure~\ref{fig:environments} depicts the four environments using the MuJoCo physics engine~\cite{MuJoCo}.

\begin{figure*}[!t]
    \centering
    \includegraphics[width=0.48\textwidth]{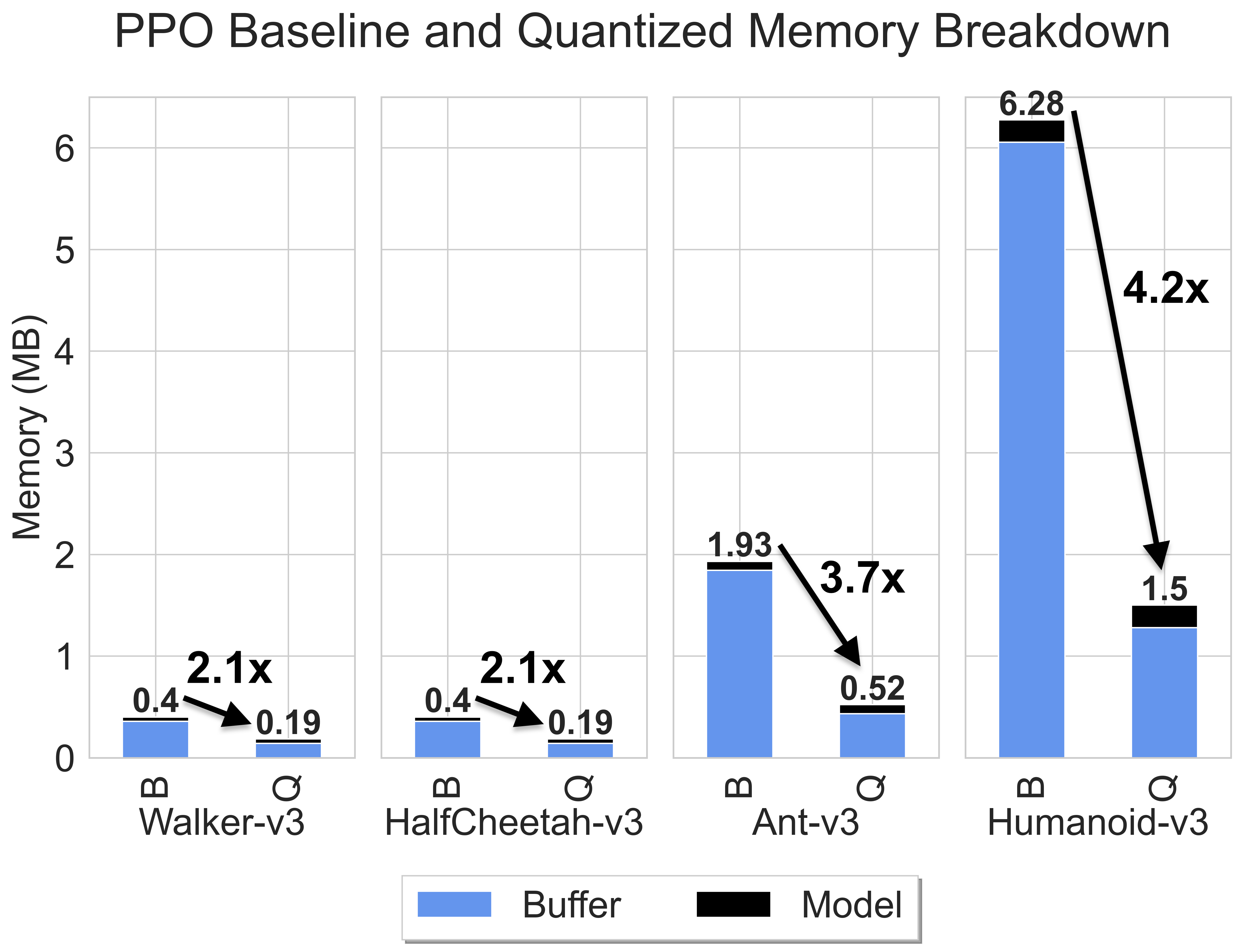}
    \includegraphics[width=0.48\textwidth]{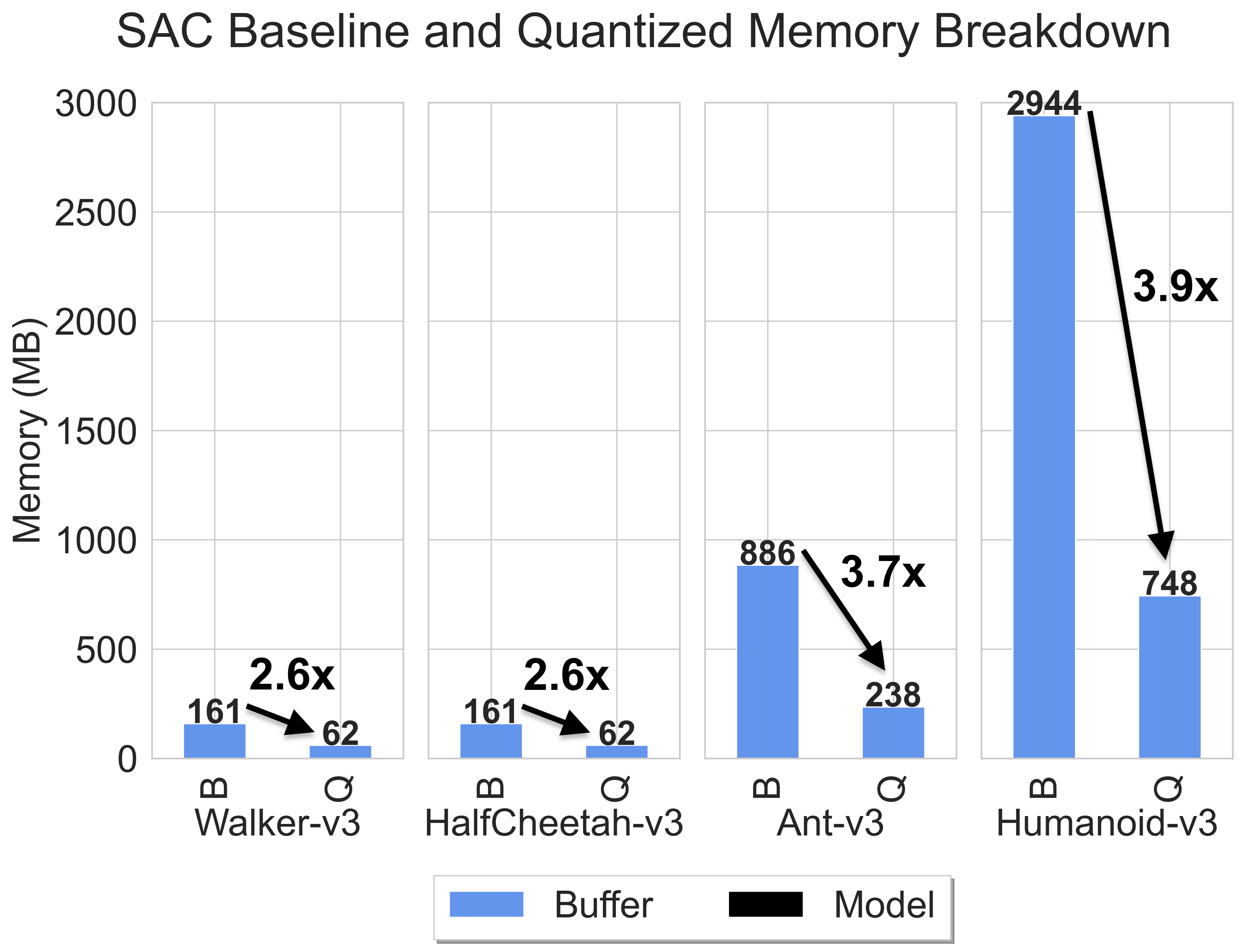}
    \caption{Memory breakdown of PPO (left) and SAC (right) both without (B) and with (Q) observation space quantization across environments. The model size is not visible in the SAC chart as it comprises less than 2\% of the total memory.}
    \label{fig:memory_breakdown}
    \vspace{-10pt}
\end{figure*}

\subsection{Implementation Details} 
Our full implementation is publicly available and can be found in our GitHub repository: \url{https://github.com/A2R-Lab/Just-Round}. 

We train all agents using the \texttt{stable-baselines3}~\cite{stable-baselines3} implementations of both PPO and SAC, modifying only the buffer logic in order to accommodate quantization of the stored observations. Unless otherwise noted, we run all experiments using the default PPO and SAC hyperparameters, only adjusting the number of total training steps.\footnote{Training done on an Intel Core i7-8700K CPU and an NVIDIA GeForce GTX 1080 Ti GPU.}

\subsection{Memory Efficiency} 
Figure~\ref{fig:memory_breakdown} compares, for all locomotion environments and for both PPO and SAC, the memory required both with and without quantizing the observations in the buffer. We breakdown the memory usage into the the policy and value/Q-value networks (shown in black) and the buffer (shown in blue). As mentioned in Section~\ref{sec:learning_bk}, the memory footprint is dominated by the buffers, rendering the model's memory footprint barely visible in the PPO plots and mostly invisible in the SAC plots, where it only accounts for less than 2\% of the total memory. 

For all experiments we set $a=127, b=-127$. For PPO we set $m=1$, and for SAC we set $m=2$ after observing that SAC required extra decimal precision to achieve optimal convergence. We hypothesize that this discrepancy between PPO and SAC is a result of the off-policy nature of SAC. While in each iteration PPO is trained using a smaller number of rollouts from the most recent policy, SAC utilizes a larger set of replay data collected from both past and present policies. This larger buffer may in turn lead to a higher chance of observation ``collisions," where two rounded tuples $(s_q, a, r_1)$ and $(s_q, a, r_2)$ present the same quantized observation and action with a different reward, leading to counterproductive training steps. 

Overall, we found that quantizing only the observations stored in the buffers reduces overall memory footprint by $2.1\times$ to $4.2\times$ for PPO and $2.6\times$ to $3.9\times$ for SAC. In the \texttt{Humanoid-v3} environment, this $3.9\times$ reduction in memory used for SAC decreases a 3GB memory requirement to just under 750MB. This space savings is significant, allowing real, memory-constrained learning systems the ability to store much larger replay buffers on-board than would be possible without quantization.

\begin{figure}[]
    \centering
    \includegraphics[width=\columnwidth]{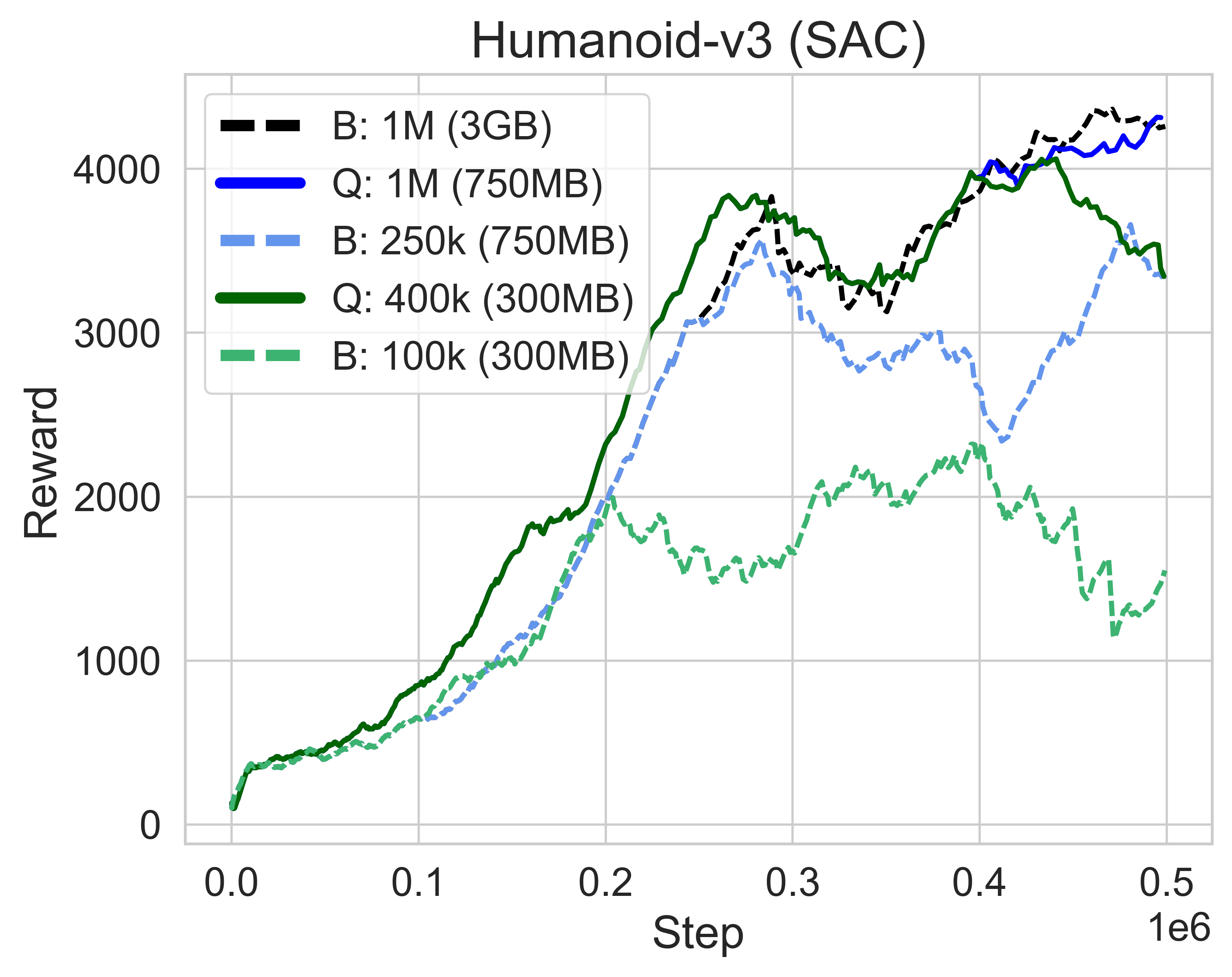}
    \caption{Learning curves of SAC applied to the $\texttt{Humanoid-v3}$ environment with (Q, solid) and without (B, dashed) quantization and with variable replay buffer sizes: 1M, 400k, 250k, 100k. Colors differentiate experiments based on memory footprint: black $\sim$3GB, dark/light blue $\sim$750MB, dark/light green $\sim$300MB. Parameters $m=2$, $a=127$, and $b=-127$ found through minimal tuning.}
    \vspace{-15pt}
    \label{fig:buffer_size}
\end{figure}

\begin{figure*}[!t]
    \vspace{2mm}
    \centering
    \includegraphics[width=0.48\textwidth]{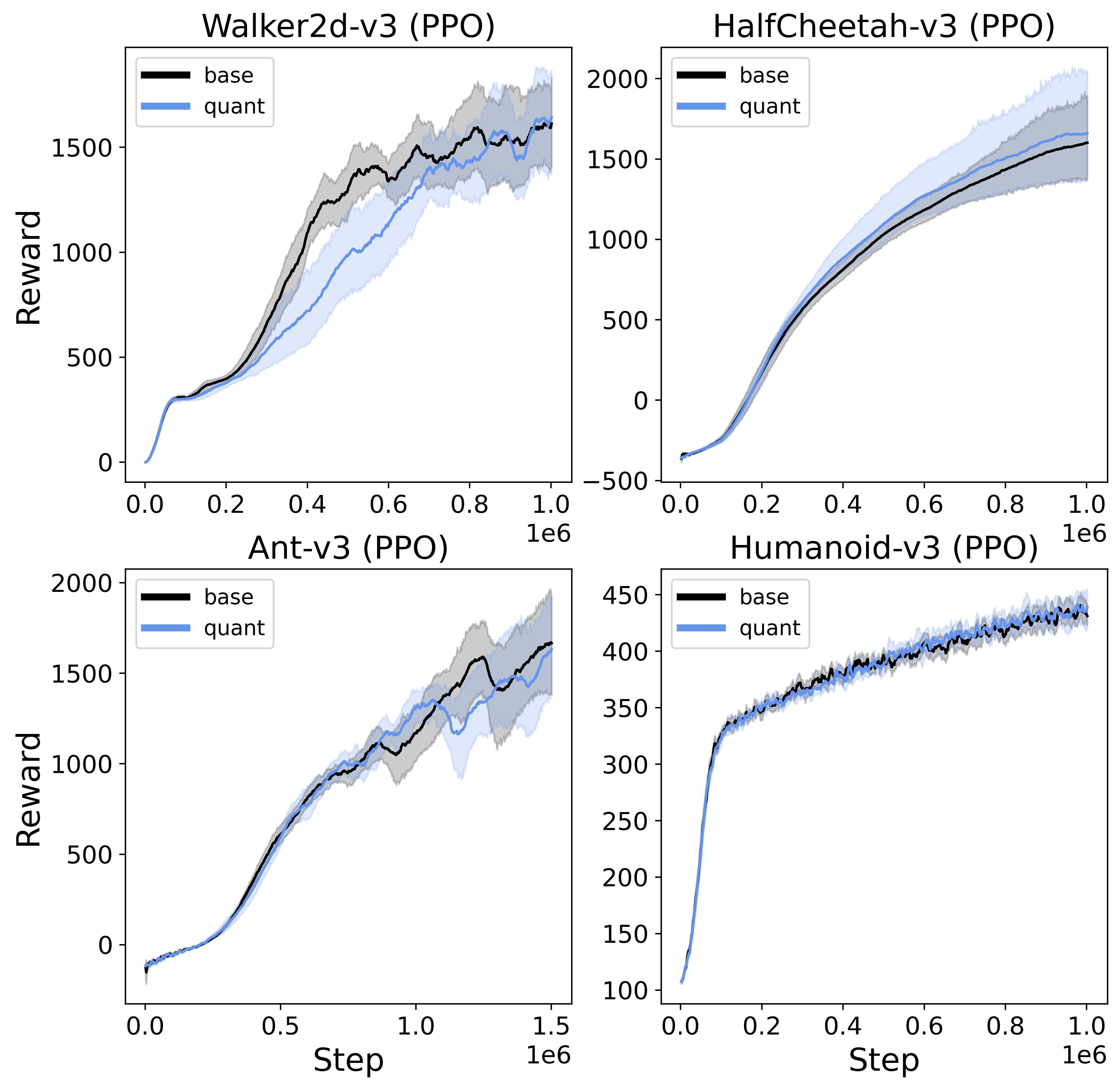}
    \includegraphics[width=0.48\textwidth]{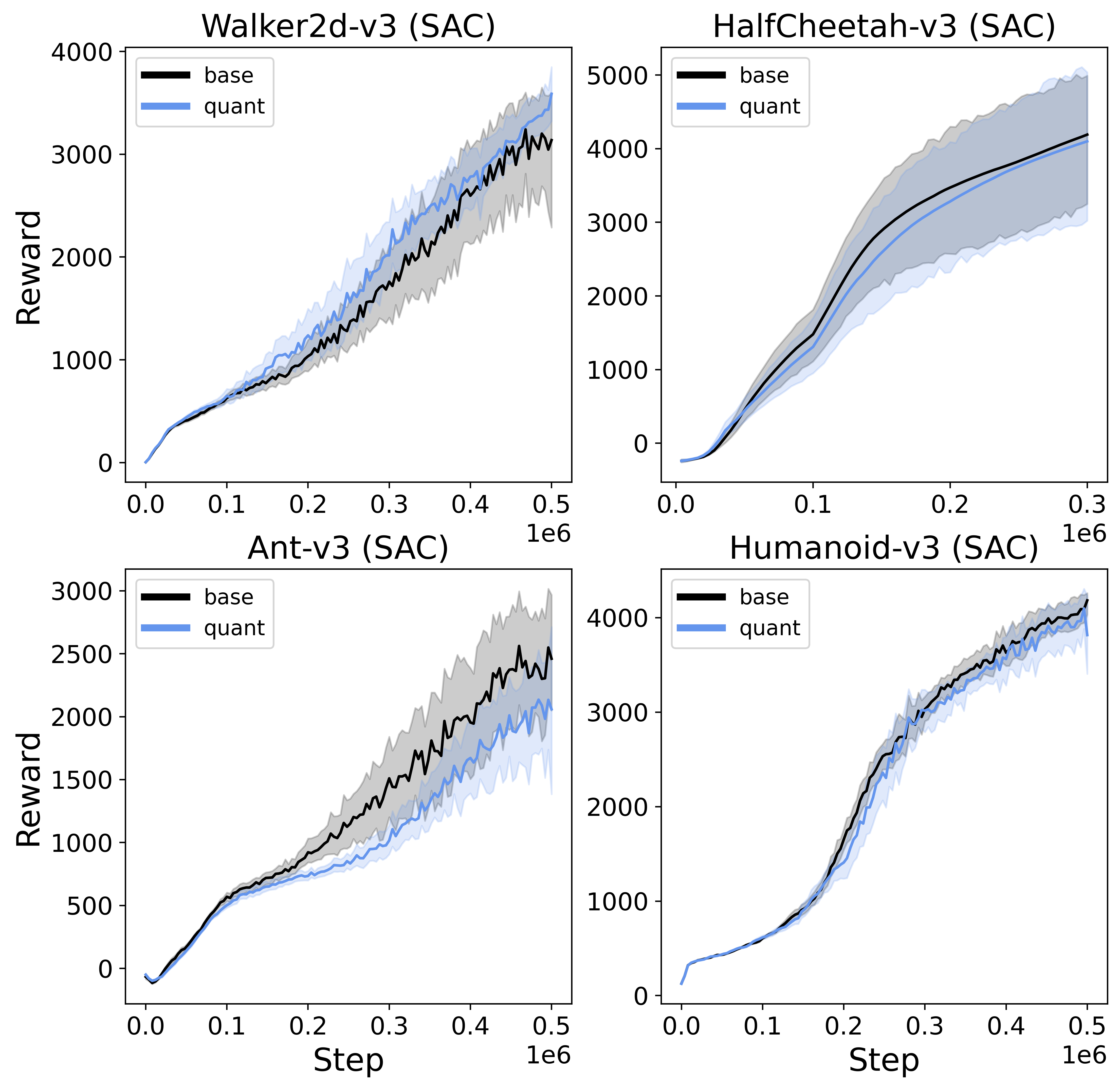}
    \caption{Learning curves of PPO (left) and SAC (right) with (blue) and without (black) quantization across four simulated locomotion environments. Rewards are averaged across ten trials and standard deviations are shaded.}
    \label{fig:learning_curves}
    \vspace{-5pt}
\end{figure*}

\subsection{Training Performance} 

Figure~\ref{fig:buffer_size} plots the learning curves for the first 500k steps of training for the $\texttt{Humanoid-v3}$ environment using both quantization and the baseline method of reducing the number of observations in the replay buffer. 
We ran all experiments with the same hyperparameters, only varying the buffer quantization and size. 
The baseline SAC agents with varied numbers of observations in the rollout buffer are plotted using dashed lines, and the quantized agents are plotted with solid lines. We group and color-code experiments by memory footprint: black corresponding to $\sim$3GB in size, dark/light blue to $\sim$750MB, and dark/light green to $\sim$300MB.

First and foremost, we find that the quantized agent with a buffer size of 1M (in solid dark blue) displays similar convergence results as the baseline 1M agent (in dashed black), while only using roughly a quarter of the memory ($\sim$750MB vs $\sim$3GB). Holding the memory footprint constant, we find that the quantized agents (solid dark blue and dark green) converge to significantly higher rewards than their unquantized counterparts (dashed light blue and light green). While stark, this difference is not entirely surprising. For more complex learning tasks, like humanoid walking, larger replay buffers are often essential in enabling performant convergence. Overall, this experiment indicates that observation-based quantization may allow us to retain large replay buffers and their associated highly performant training while reducing memory costs by as much as $4.2\times$.




\subsection{Training Convergence and Speed}
To validate the generalizability of this result we show the learning curves of PPO and SAC across four different simulated locomotion environments in Figure~\ref{fig:learning_curves}. These figures report the average of ten random seeds and also display the standard deviation with the accompanying shaded regions. 

We find that convergence is not substantially affected by the introduction of observation-based quantization. All final rewards of both quantized and unquantized agents are within a standard deviation of each other, and in some cases the quantized models are able to train a policy with a higher total reward. Furthermore, we show in Table~\ref{table:wall_time} that the overhead from performing quantization also does not significantly impact overall training times and in some cases actually speeds up overall training through reductions in memory access overheads.

\begin{table}[h!]
\centering
\bgroup
\def\arraystretch{1.2} 
\begin{tabular}{ |c|c|c|c|c|c| } 
 \hline
  & & \texttt{Walker2d} & \texttt{HalfCheetah} & \texttt{Ant} & \texttt{Humanoid}\\
 \hline
 \multirow{3}{2em}{PPO} & B    & 1.66 & 1.77 & 2.16 & 1.78 \\ 
                        & Q & 1.61  & 1.76 & 2.17 & 1.87 \\
                        & T & 0.97$\times$  & 0.99$\times$ & 1.00$\times$ & 1.05$\times$ \\
 \hline
 \multirow{3}{2em}{SAC} & B    & 8.38 & 8.49 & 9.00 & 9.83 \\ 
                        & Q & 8.41 & 8.44 & 8.97 & 9.91 \\ 
                        & T & 1.00$\times$  & 0.99$\times$ & 1.00$\times$ & 1.01$\times$ \\
 \hline
\end{tabular}
\egroup
\caption{Wall clock time in milliseconds per step of PPO and SAC with (Q) and without (B) quantization as well as the relative total time (T) of training with quantization.}
\vspace{-15pt}
\label{table:wall_time}
\end{table}

\section{CONCLUSION AND FUTURE WORK}
This paper presents observation space quantization, a method for reducing the overall memory requirements of DRL without impacting training performance.

We evaluated the overall memory footprint and learning performance for four simulated robotic locomotion tasks using two state-of-the-art on- and off- policy DRL algorithms. 
We found that applying observation space quantization reduced the effective memory footprint of DRL training by as much as $4.2\times$, while preserving both training convergence and speed.
We believe that this approach can be a powerful way to reduce the overall memory consumption of DRL. 

While we admit that not every learning-based solution requires memory compression (especially if learning is being done on a powerful workstation with ample storage), if we are to achieve lifelong, practical learning on the edge, curbing memory usage is a must.

In future work, we hope to compare our approach to other forms of reduced precision (e.g., half-precision floating point~\cite{Bjork21a}). We also hope to deploy our approach onto physical robot hardware and test it in the context of real-world edge RL. 
Finally, we hope to apply our approach in the context of tiny robot learning~\cite{Neuman22TinyRobotLearning} and help usher in a new era of ubiquitous edge RL.

   
\section*{Appendix A: Learning Hyperparameters}
All experiments were carried out using the default $\texttt{stable-baselines3}$ learning parameters for PPO and SAC unless otherwise noted. The full hyperparameter list is given in Table~\ref{table:hyperparameters}. Both PPO and SAC used two-layered feed-forward networks with 64 nodes per layer.

\begin{table}[h!]
\centering
\bgroup
\def\arraystretch{1.2} 
\begin{tabular}{ |l|l|c| }
  \hline
  & Parameter & Value \\
  \hline
   \multirow{8}{2em}{PPO} & MDP steps per update& 2048\\
   & batch size & 64\\
   & learning rate & 0.0003\\
   & gamma & 0.99\\
   & $\lambda$ for GAE & 0.95\\
   & value function cost & 0.5\\
   & entropy cost & 0.0\\
   & max gradient norm & 0.5\\
   & cliprange & 0.2\\
 \hline
   \multirow{7}{2em}{SAC} & learning rate & 0.0003\\
   & buffer size & 1000000\\
   & entropy coefficient & auto\\
   & MDP steps between updates & 1\\
   & batch size & 256\\
   & tau & 0.005\\
   & gamma & 0.99\\
   & gradient steps per update & 1\\
 \hline
\end{tabular}
\egroup
\caption{Learning hyperparameters}
\label{table:hyperparameters}
\vspace{-10pt}
\end{table}







\bibliographystyle{inc/IEEEtran}
\bibliography{inc/IEEEabrv,inc/main}

\end{document}